\documentclass[11pt]{article}
\usepackage{hyperref}
\usepackage{url}
\usepackage{amsmath}
\usepackage[pdftex]{graphicx}
\usepackage{cite}
\usepackage{color}
\usepackage[margin=1in]{geometry}

\begin{document}

\title{\textbf{Artificial Intelligence Enables Real-Time and Intuitive Control of Prostheses via Nerve Interface}}

\author{Diu Khue Luu$^{*1}$,
			Anh Tuan Nguyen$^{*1, 5}$,
			Ming Jiang$^{2}$,
			Markus W. Drealan$^{1}$,
			Jian Xu$^{1}$,\\
			Tong Wu$^{1}$,
			Wing-kin Tam$^{1}$,
			Wenfeng Zhao$^{1}$,
			Brian Z. H. Lim$^{1}$,
			Cynthia K. Overstreet$^{4}$,\\
			Qi Zhao$^{2}$,
			Jonathan Cheng$^{3}$,
			Edward W. Keefer$^{4, 5}$,
			Zhi Yang$^{1, 5}$}
\date{	\raggedright $^{*}$ Co-first authors \\
			\raggedright $^{1}$ Biomedical Engineering, University of Minnesota, Minneapolis, MN, USA \\
			\raggedright $^{2}$ Computer Science and Engineering, University of Minnesota, Minneapolis, MN, USA \\
			\raggedright $^{3}$ Plastic Surgery, University of Texas Southwestern Medical Center, Dallas, TX, USA \\
			\raggedright $^{4}$ Nerves Incorporated, Dallas, TX, USA \\
			\raggedright $^{5}$ Fasikl Incorporated, Minneapolis, MN, USA \\
			\raggedright Correspondence: D. K. Luu (Email: luu00009@umn.edu) and A. T. Nguyen (Email: nguy2833@umn.edu)\\}
\maketitle

\maketitle

\begin{abstract}

\textit{Objective}: The next generation prosthetic hand that moves and feels like a real hand requires a robust neural interconnection between the human minds and machines.
\textit{Methods}: Here we present a neuroprosthetic system to demonstrate that principle by employing an artificial intelligence (AI) agent to translate the amputee’s movement intent through a peripheral nerve interface. The AI agent is designed based on the recurrent neural network (RNN) and could simultaneously decode six degree-of-freedom (DOF) from multichannel nerve data in real-time. The decoder's performance is characterized in motor decoding experiments with three human amputees.
\textit{Results}: First, we show the AI agent enables amputees to intuitively control a prosthetic hand with individual finger and wrist movements up to 97-98\% accuracy. Second, we demonstrate the AI agent's real-time performance by measuring the reaction time and information throughput in a hand gesture matching task. Third, we investigate the AI agent's long-term uses and show the decoder's robust predictive performance over a 16-month implant duration.
\textit{Conclusion \& significance}: Our study demonstrates the potential of AI-enabled nerve technology, underling the next generation of dexterous and intuitive prosthetic hands.
\end{abstract}


\section{Introduction}

The number of upper-limb amputees in the U.S. has been escalating during the past few decades. From a population of 541,000 in the year 2005, the amputee population is expected to exceed 1 million by the year 2050 \cite{2008_ZieglerGraham}. While many upper-limb prostheses with individually motorized joints and fingers have become commercially available, limitations of the control scheme have not enabled significant improvements of amputee's well-being. Numerous advanced motor decoders are being developed to fulfill this missing link. There are several approaches that can be generally categorized based on the inputs signals: myoelectric-based (electromyography, EMG), brain/cortical recording, and peripheral nerve (electroneurography, ENG).

EMG-based system is the most popular approach because it is noninvasive, relatively easy to obtain, and has been proven to hold useful information for motor decoding \cite{2013_Amsuss, 2012_Jiang, 2012_Fougner}. EMG prosthesis control has a long history dated back to the 1960s \cite{1960_Kobrinskiy, 1965_Popov}. In recent years, advances in machine learning have further pushed the accuracy boundary of EMG-based motor decoders. The current state-of-the-art for an offline classification task is using support-vector machine (SVM) to decode EMG signals, which yields an accuracy up to 99\% for a 4-class problem \cite{2012_Alkan}. However, \cite{2015_Ortiz} shows that good offline decoding outcomes do not necessarily translate to real-time performance. In \cite{2021_Souza}, decoding EMG signals with multilayer perceptron (MLP) yields an accuracy of only 82\% for a 12-class problem on transradial amputees. Furthermore, a significant limitation of most commercial EMG-based control is that amputees must learn to contract their residual muscles in predefined patterns to map into different hand or wrist motions. This control scheme is inherently unnatural and requires much user training. EMG signals are also susceptible to electrode repositioning and motion artifacts \cite{2016_Cordella}.

A direct neural interface that decodes true movement intents from the brain or peripheral nerve signals promises an intuitive solution for prosthesis control, yet they come with their own set of challenges. Although a brain implant offers the most comprehensive human-machine interconnection, it is invasive and carries significant risks of neural tissue damage in long-term uses \cite{2016_Flesher, 2018_Collinger, 2010_Schalk}. On the other hand, a peripheral nerve interface like one reported in our previous work \cite{2018_Xu_Nerve, 2020_Nguyen_Scorpius, 2021_Nguyen_RXF} and others \cite{2015_Pasquina, 2015_Schiefer, 2015_Saal, 2016_Lachapelle,2016_Davis, 2017_Vu, 2017_Wendelken, 2020_Cracchiolo, 2020_Vu, 2019_Wolf} is less invasive while still providing sufficient motor control signals with simultaneous somatosensory neuro-feedback. These interfaces aim to enable intuitive prosthesis control purely by thoughts and achieve a natural user experience, which is crucial for the amputee to take full advantage of near-anatomic prosthetic hands like the LUKE Arm \cite{2014a_Resnik, 2014b_Resnik} with at least 10 degrees of freedom (DOF) of movement. Decoding motor intent from the peripheral nerves without input from the residual muscles also allows the system to be used by a larger population of amputees with various amputation levels. Despite the promises, one of the major challenges for materializing this technology into clinical prostheses is that the interface generates a large amount of high-dimensional data, which must be efficiently translated to prosthesis movements in real-time.

Artificial intelligence (AI) based on deep learning has emerged as the most prominent approach to leverage this challenge \cite{2012_Sussillo, 2016_Atzori,2018_George, 2019_Alazrai, 2019_Dantas, 2020b_George, 2020_Nguyen_Scorpius, 2021_Luu_Frontiers}. Our previous works \cite{2020_Nguyen_Scorpius, 2021_Luu_Frontiers} demonstrate that AI neural decoders based on the convolutional neural network (CNN) and recurrent neural network (RNN) architecture excel over other machine learning techniques, including SVM, random forest (RF), and MLP in both classification and regression tasks. Furthermore, we show that a deep learning AI can be efficiently deployed on a portable, self-contained system thanks to advances in edge computing \cite{2021_Nguyen_Draco}.

Here we focus on characterizing the motor decoding performance of an AI agent based on the RNN architecture. This is done through multiple motor decoding experiments with three human amputees who receive microelectrodes implants from 6 to 16 months. First, we show the AI agent can simultaneously decode the subject's motor intents with six DOF, including individual fingers flexing and wrist pronation. The prediction accuracy ranges from 85-93\% with one subject to 97-98\% with another. Second, we design a hand gesture matching task derived from the mental chronometry test to quantify the AI agent's real-time response \cite{1978_Posner, 2006_Jensen}. The task measures the time taken for the subject to compose a targeted gesture in random order. The results show a median reaction time of 0.81 sec, corresponding to an information throughput of 6.09 bps (365.4 bpm) at 99\% accuracy. Third, we study the AI agent's long-term use by evaluating the prediction performance at various time points over a 16-month implantation duration. We show that the nerve data's signal-to-noise (SNR), reaction time, and information throughput remain robust throughout the course of the experiment, while the AI agent only requires fine-tuning every few months. In fact, the best motor decoding performance is observed in the last experiment session before the explant surgery. 

The rest of the manuscript is organized as follows. Section \ref{Sec_Methods} describes the human subjects, the AI neural decoder, and the experiment setup. Section \ref{Sec_Results} presents the experiment results and findings. Section \ref{Sec_Discussion} provides discussions about the results and future directions. Section \ref{Sec_Conclusion} concludes the paper.

\section{Methods}
\label{Sec_Methods}

\subsection{Human subjects and experiment protocol}

The human experiment is a part of the clinical trial DExterous Hand Control Through Fascicular Targeting (DEFT)  identifier No. NCT02994160\footnote{\url{https://clinicaltrials.gov/ct2/show/NCT02994160}}, which is sponsored by the DARPA Biological Technologies Office as part of the Hand Proprioception and Touch Interfaces (HAPTIX) program.

The human experiment protocols are reviewed and approved by the Institutional Review Board (IRB) at the University of Minnesota (UMN) and the University of Texas Southwestern Medical Center (UTSW). The amputees voluntarily participate in our study and are informed of the methods, aims, benefits, and potential risks of the experiments before signing the Informed Consent. Patient safety and data privacy are overseen by the Data and Safety Monitoring Committee (DSMC) at UTSW. The implantation, initial testing, and post-operative care are performed at UTSW by Dr. Cheng and Dr. Keefer, while motor decoding experiments are performed at UMN by Dr. Yang's lab. The clinical team travels with the subject in each experiment session. The subjects also complete the Publicity Agreements where they agree to be publicly identified, including showing their face.
\begin{table*}[ht]
\centering
\caption{\normalsize Summary of human subjects}
\includegraphics[width=0.9\textwidth]{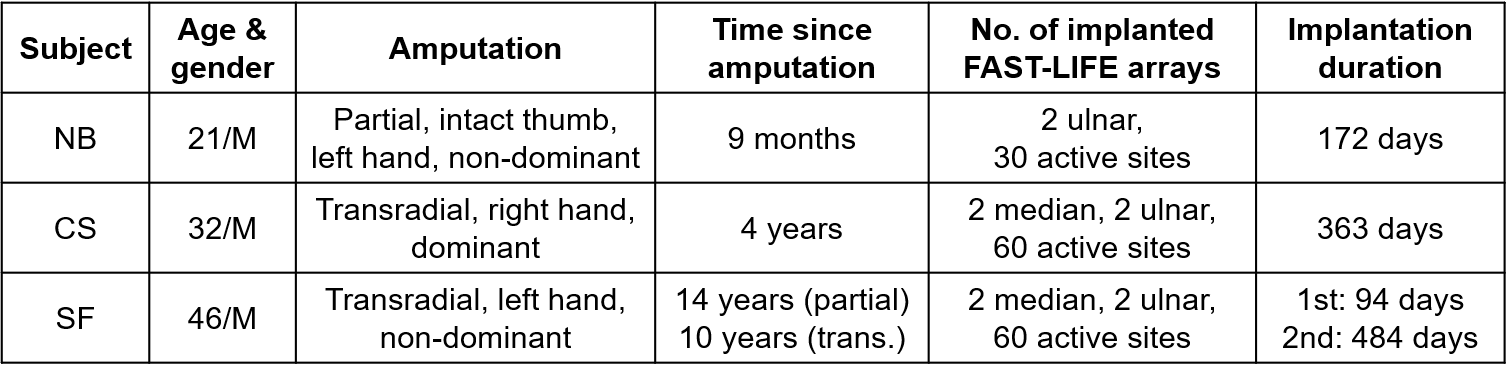}
\label{Table_Patient}
\end{table*}

Table \ref{Table_Patient} summarizes the three human amputees, who belong to a subset of a larger study described in \cite{2017_Cheng, 2019_Overstreet, 2021_Cheng}, aiming to create a robust nerve interface with both recording and stimulation capabilities. Each subject receive 2-4 fascicle-specific targeting of longitudinal intrafascicular electrodes (FAST-LIFE) microelectrode implants targeting individual fascicles in the median and ulnar nerve for a span of 3 to 16 months. The electrode configuration and implant duration depend on the current IRB protocol and the individual subject's physiology during the study.

In particular, Subject NB has two implants in the ulnar nerve, while Subject CS and SF have four implants in both the median and ulnar nerve. Subject NB's participation was interrupted due to an unrelated motorcycle incident, thus the microelectrodes were explanted early for safety reasons. Subject CS and SF participated in the full experiment duration. Especially, Subject SF volunteered for two separate implant courses. After the first course of 3 months, all four microelectrode arrays were explanted from the Subject SF’s arm according to the existing IRB protocol. However, Subject SF reported positive experiences with the clinical study’s outcomes and voluntarily participated in an additional implant course. About a year later, a subsequent IRB modification was made, and Subject SF had four new microelectrode arrays implanted. His second course ultimately lasted for 16 months, which allows us to characterize the long-term performance of the proposed nerve interface and motor decoder. The results reported in Section III. B, C, including Figure \ref{Fig_Trajectory}, \ref{Fig_ReactionTime}, \ref{Fig_Persistance} and Table \ref{Table_Info} are derived from Subject SF's second course.

\subsection{Deep learning-based AI neural decoder}
\begin{figure*}[ht]
\centering
\includegraphics[trim=0 0 0 0, clip=true, width=0.8\textwidth]{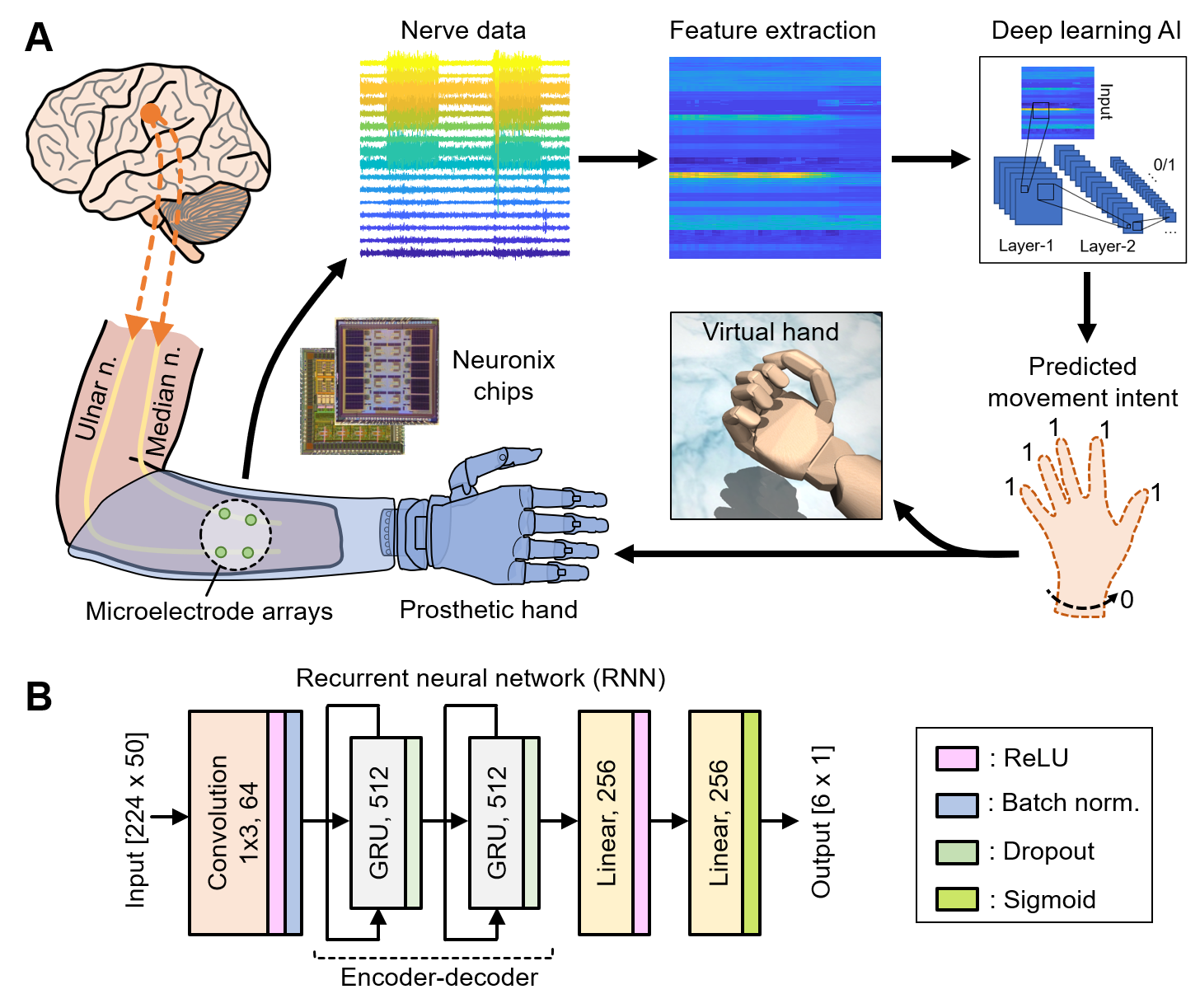}
\caption{(A) Overview of the AI neural decoder and signal processing paradigm. Nerve data are acquired from the subject's amputated arm by Neuronix neural interface chips, followed by feature extraction. The deep learning AI then uses feature data to predict the subject's intent of moving several DOF simultaneously. The predictions are mapped to movements of a virtual hand or a prosthetic hand in real-time. (B) Design of the deep learning AI based on the recurrent neural network (RNN) architecture.}
\label{Fig_Decoder}
\end{figure*}

Figure \ref{Fig_Decoder}(A) shows the overview of the AI and signal processing paradigm. Nerve data are acquired with our Scorpius neuromodulation system powered by high-performance Neuronix neural interface chips with both neural recorders and stimulators. The Scorpius device and Neuronix chips specifications are detailed in \cite{2020_Nguyen_Scorpius}. Up to 16 channels are recorded, which belong to a subset of the active sites. Most channels are from the nerve fascicle with the majority of motor fibers. We select sites that have high SNR and strong voluntary compound action potentials (vCAP) activities during finger flexing. The signal's strength $P_{\text{dB}}$ and SNR are estimated as follow:
\begin{align}
	P_{\text{dB}} &= 10 \log_{10} ( 1/n \Sigma_{i=1}^n V[i]^2 ) \\
	\text{[SNR]} &= P_{\text{dB, flex}} / P_{\text{dB, rest}}
\end{align}
Raw nerve signals are acquired at 10 kHz sampling rate, downsampled to 5kHz, and then filtered in the bandwidth of 25-600 Hz. The chosen bandwidth contains most of the motor control signal's power, comprising primarily of vCAP. 

We then perform feature extraction on the filtered signals with a 100 msec sliding window and 20 msec step. 14 features are extracted from each channel as detailed in \cite{2021_Luu_Frontiers, 2021_Nguyen_Draco}. They include zero crossing (ZC), slope sign changes (SSC), waveform length (WL), Wilson amplitude (WA), mean absolute (MAB), mean square (MSQ), root mean square (RMS), V-order 3 (V3), log detector (LD), difference absolute standard deviation (DABS), maximum fractal length (MFL), myopulse percentage rate (MPR), mean absolute value slope (MAVS), and weighted mean absolute (WMA).

At any given time, the past 1 sec of nerve data is used by the AI agent to predict the current motor intents. A 2D data representation with dimensions [16 channels] $\times$ [14 features] $\times$ [50 time-series] = [224 $\times$ 50] is compiled as the input to the deep learning AI agent. The AI agent simultaneously classifies the movement of 6 DOF: 5 for individual finger flexing and 1 for wrist pronation. Each DOF is assigned class-0 for ``resting'' or class-1 for ``flexing''. Depending on the hardware capability, the prediction rate (frame per second) can be set from 5-50 Hz. The prediction output is mapped to the movement of either a virtual hand (MuJoCo, \url{mujoco.org}) or a physical prosthetic hand with individually actuated fingers. The prosthetic hand typically has a slower response than the virtual hand due to mechanical constraints. 

Figure \ref{Fig_Decoder}(B) shows deep learning AI's design based on the RNN architecture. The model is implemented in the PyTorch framework (\url{pytorch.org}) and utilizes standard layers, including convolution, gated recurrent units (GRU), linear/fully-connected, rectified linear unit (ReLU), batch normalization, 50\% dropout, and sigmoid. In total, the model consists of 1.6 million trainable parameters. This design is relatively shallow because the model must be able to be deployed in a portable edge computing device like the NVIDIA Jetson Nano \cite{2021_Nguyen_Draco}. 
\begin{figure*}[p]
\centering
\includegraphics[trim=0 0 0 0, clip=true, width=1\textwidth]{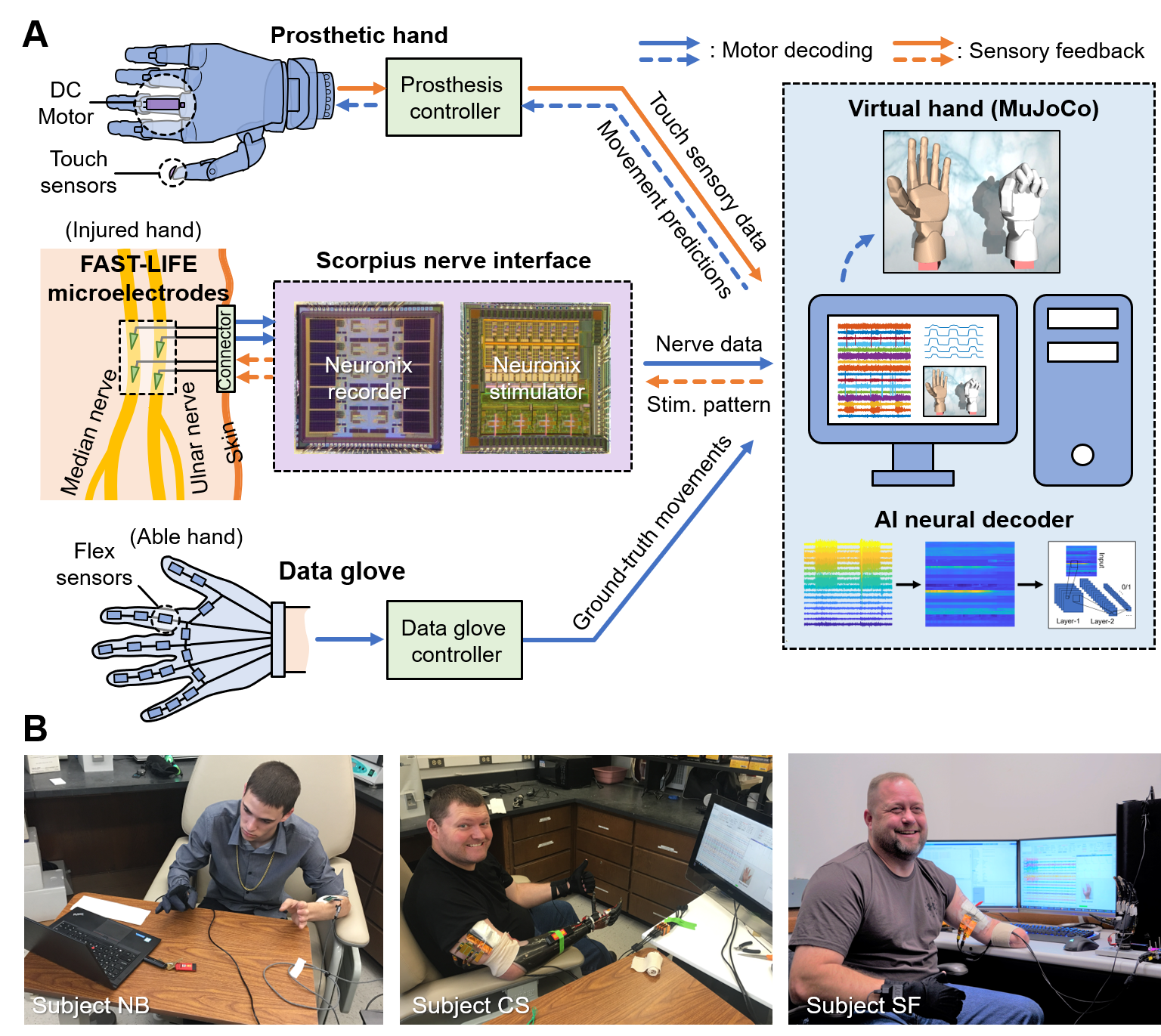}
\caption{(A) Overview of the experiment setup with both neural recording and stimulation capabilities. (Blue path) The motor decoding dataset is obtained via a mirrored bilateral paradigm. Nerve data and ground-truth movements are simultaneously acquired from the injured and able hand, respectively. All signal processing, neural decoding, and real-time displaying are done on a desktop PC. Movement predictions can be mapped to a prosthetic hand or a virtual hand. (Orange path) The setup also includes components like touch sensors and neurostimulators for somatosensory restoration as detailed in \cite{2019_Overstreet, 2021_Cheng}. (C) Photos of three subjects in an experiment session.}
\label{Fig_ExpSetup}
\end{figure*}

We use the Adam optimizer \cite{2014_Kingma} for training the model with the default parameters $\beta_1 = 0.99$, $\beta_2 = 0.999$, weight decay regularization $L_2 = 10^{-5}$, mini-batch of 64, initial learning rate of $10^{-4}$, and maximum number of epoch of 5. The learning rate is reduced by a factor of 10 when the training loss stops improving for two consecutive epochs. The final activation function is the sigmoid function. Therefore, the predicted probabilities for each DOF fall in the range [0, 1]. We choose a threshold of 0.5 for all DOFs to classify whether a DOF is flexing. Throughout the course of the experiment, the model could be fine-tuned by adding new data for training without altering its architecture regarding the hidden layers' types, numbers, sequences, and activation functions. Model training, validation, fine-tuning and online motor decoding experiments are performed on a desktop PC with an Intel Core i7-8086K CPU and an NVIDIA GTX 1080Ti GPU.

In addition, we also utilize a stochastic approach to further optimize the AI agent's performance. Because the neural decoding problem is not convex, every time the model is trained with a random initial seed, it would converge to a different local minimum. By retraining it numerous times with different random seeds and only keeping the solution with the best overall accuracy across all DOF, we increase the chance to obtain the global minimum. Each training iteration takes about 30-60 sec to complete depending on the dataset's size. Our experiments are several months apart. Before any experiment, we first train the AI agent with old data for a few hours and evaluate the model's persistence in real-time motor decoding experiments. We then record new nerve data, retrain the model for a few minutes, and repeat the real-time experiments to measure the up-to-date performances.

\subsection{Motor decoding experiment setup}

Figure \ref{Fig_ExpSetup}(A, B) show an overview of the experiment setup and photos of three subjects in a session. The system has both neural recording and stimulation capabilities for motor decoding and somatosensation feedback. Data acquisition, real-time processing, decoding, display, and storage are made with a desktop PC. The dataset is obtained via a mirrored bilateral paradigm. In each experiment session, the subject repeatedly makes a hand gesture 10 times with both hands. Nerve data are acquired from the injured/phantom hand with the Scorpius nerve interface, while ground-truth labels are simultaneously captured from the able hand with a data glove. The same setup without the data glove is used for real-time motor decoding. The AI agent's prediction can be both mapped to the virtual hand and sent to the prosthetic hand by Bluetooth. The setup also includes components like touch sensors and neurostimulators for somatosensory restoration experiments as detailed in \cite{2021_Nguyen_RXF, 2021_Cheng}.
\begin{table}[ht]
\centering
\caption{\normalsize Summary of motor decoding datasets}
\includegraphics[trim=5 0 0 0, clip=true, width=0.6\textwidth]{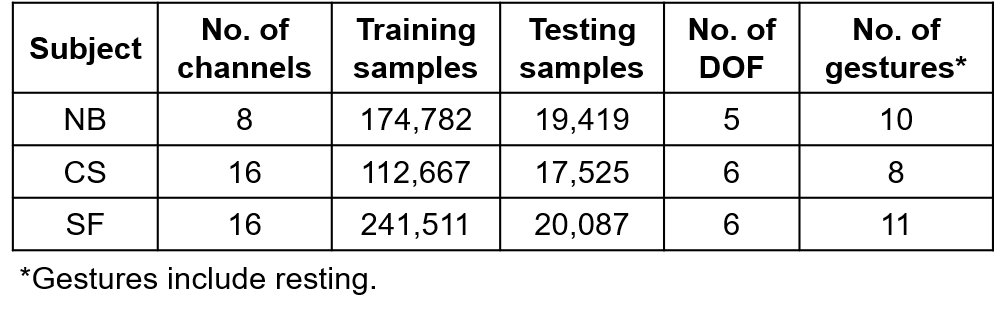}
\label{Table_TrainTest}
\end{table}

Table \ref{Table_TrainTest} summarizes typical motor decoding datasets collected in one day. They are the particular sets that would be used for evaluating classification performance in Section \ref{Sec_Classification}. We collect similar datasets on different days a few months apart to characterize the nerve interface's long-term persistence. Each nerve dataset is specific to each subject. The list of gestures includes resting (000000), individual fingers flexing (10000, 010000,...) and various combinations like fist/grip (111110), pinch (110000), etc. that are repeated in an alternate order. We record as many sessions as time permits. The last recording session, which contains the most up-to-date nerve data, is always used for validation while the remaining are used for training. The data collection requires approximately 1-2 h but could take longer with intermittent breaks.  

The classification performance is evaluated using standard metrics including sensitivity/true positive rate (TPR), specificity/true negative rate (TNR), and balanced accuracy derived from true-positive (TP), true-negative (TN), false-positive (FP), and false-negative (FN) as follows:
\begin{align}
	\text{[TPR/sensitivity]} &= \text{TP}/(\text{TP}+\text{FN}) \\
	\text{[TNR/specificity]} &= \text{TN}/(\text{TN}+\text{FP}) \\
	\text{[Balanced accuracy]} &= (\text{TPR}+\text{TNR})/2 \\
	\text{[Prediction error]} &= 1 - \text{[Balanced accuracy]}
\end{align}

\subsection{Hand gesture matching task}
\label{Hand_Matching}
\begin{figure*}[p]
\centering
\includegraphics[trim=0 0 0 0, clip=true, width=0.8\textwidth]{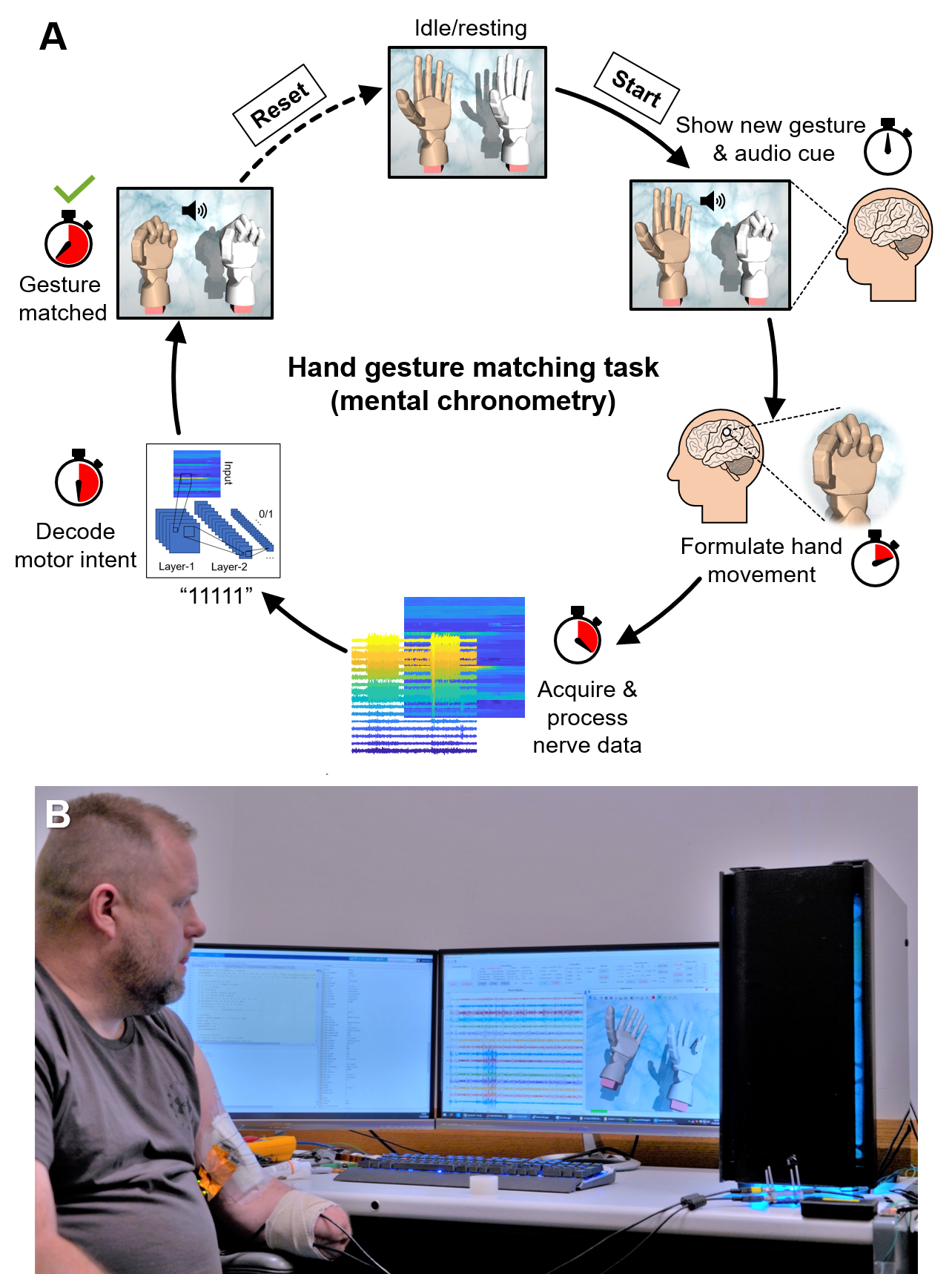}
\caption{(A) Schematic of the hand gesture matching task to measure the end-to-end reaction time (mental chronometry) and the information throughput of the entire nerve interface and motor decoding. In each trial, the subject is shown a random hand gesture, which he attempts to match with the AI neural decoder running in real-time. (B) Photo of Subject SF performing the task. Real-time nerve data and current AI's prediction can be seen on the monitor.}
\label{Fig_MatchingTask}
\end{figure*}

Figure \ref{Fig_MatchingTask}(A) shows a schematic of the hand gesture matching task derived from the mental chronometry test. It is designed to measure the end-to-end reaction time and information throughput of the entire nerve interface and motor decoding. Figure \ref{Fig_MatchingTask}(B) shows a photo of Subject SF performing the matching task with the nerve data and virtual hands visible on the monitor. The left virtual hand is the real-time AI agent's prediction that is always running. The right hand is the targeted gesture that needs to be matched. 

In each trial, the subject is shown a random hand gesture with an audio cue. The subject attempts to compose the desired gesture from the resting position with the AI decoder running in real-time, producing a movement prediction every 100 msec (10 Hz). The subject is allowed to repeat the movement multiple times until the target is matched. A success is registered at the first occurrence of the prediction outcome, where all 6 DOF must match the target. The reaction time is calculated from the moment a new target is shown till a successful match is achieved. We choose a cut-off of 3 sec, where a trial is considered unsuccessful if it fails to reach the target. The resulted reaction time and success rate reflect the end-to-end responsiveness of the entire nerve interface and AI decoder from the amputee's point of view in real-world uses. It includes information processing time of both the human mind and machine:
\begin{equation} 
\label{Eq_ReactionTime}
\begin{split}
	[\text{Reaction time}] &= [\text{Sensory acquisition (human)}] \\
	& + [\text{Cortical processing (human)}] \\
	&+ [\text{Nerve data processing (machine)}]  \\
	&+ [\text{Motor decoding (machine)}] \\
\end{split}
\end{equation}

Furthermore, the experiment allows calculating the information throughput, also known as the information transfer rate (ITR), of the nerve decoding system as follows:
\begin{align}
	[\text{Info. thoughput}] = [\text{Success rate}] \times \frac{[\text{Info. per trial}]}{[\text{Reaction time}]}
\end{align}	

The amount of information per trial is computed using Shannon's entropy formula \cite{1948_Shannon}. Subject SF is asked to match 9 targeted hand gestures, including resting, individual finger flex, fist, index pinch, and wrist pronation that are randomly assigned in each trial. The calculation includes the resting gesture even though no finger is moving because resting is a conscious selection by the user and must be accurately decoded by the AI agent. Resting accounts for 50\% of the AI agent prediction outcomes in the dataset. The remaining outcomes are randomly distributed among the other eight different hand gestures in a fair chance. Each trial in the matching task always includes two conscious gestures: resting at the beginning and one of the other gestures at the end. Subsequently, the information per trial in a matching task with resting and eight other gestures is:
\begin{equation} 
\begin{split}
	[&\text{Info. per trial}] \\
	&= 2 \cdot [-p_{rest} \log_2(p_{rest}) - 8 \cdot p_{other} \log_2(p_{other})] \\
	&= 2 \cdot [ -0.5 \cdot \log_2(0.5) - 8 \cdot (0.5/8) \cdot \log_2(0.5/8)] \\
	&= 5 \text{ bits}
\end{split}
\end{equation}

\section{Results}
\label{Sec_Results}

\subsection{Dexterous and intuitive decoding of motor intents}
\label{Sec_Classification}
\begin{figure*}[ht]
\centering
\includegraphics[trim=0 0 0 0, clip=true, width=1\textwidth]{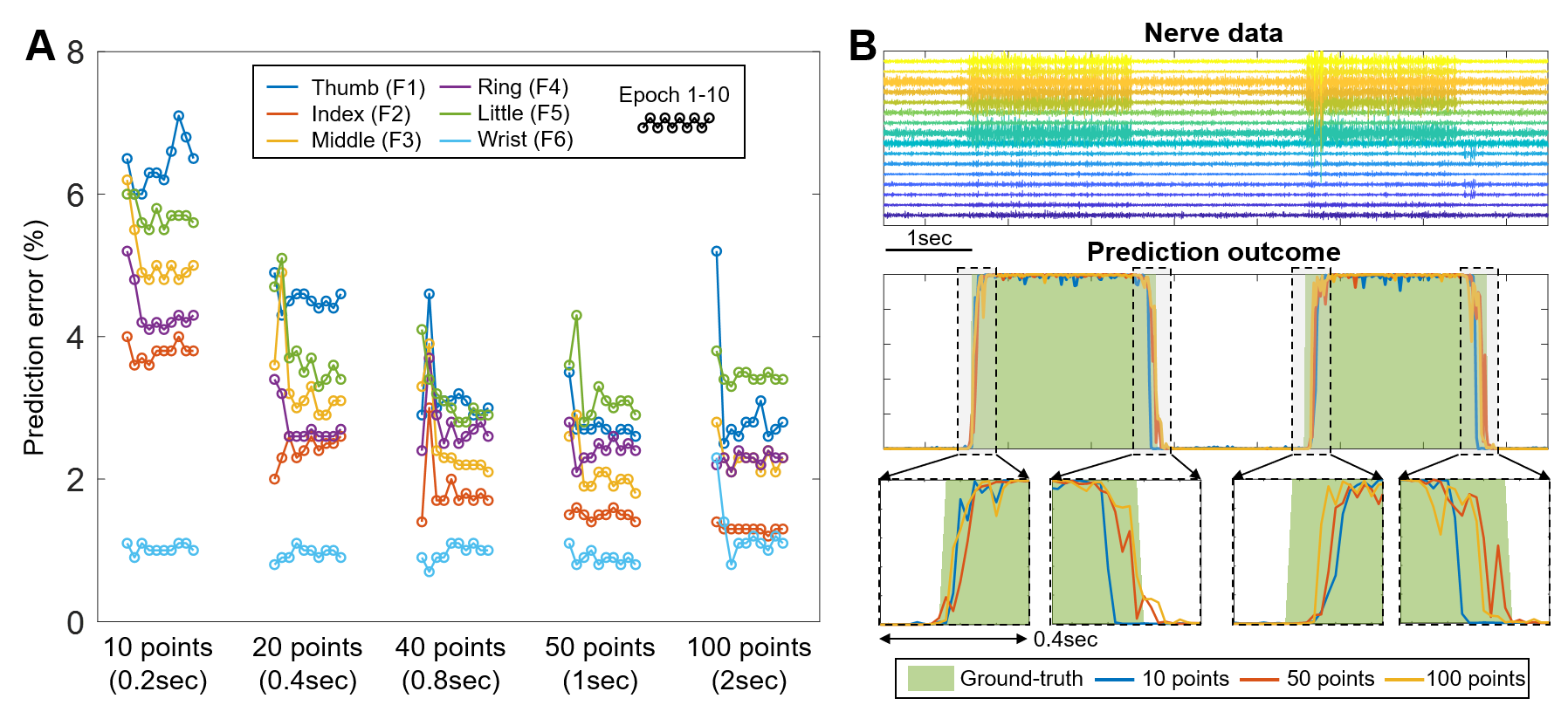}
\caption{(A) Impact of input data length on the prediction accuracy. The results suggest that decoding 1 sec of past nerve data optimally balances accuracy, complexity, and practicality. (B) Real-time prediction outcomes with different input data lengths. The results indicate that there is no significant time latency when using a longer segment of past data to decode hand movements.}
\label{Fig_PastData}
\end{figure*}

The input data segment's length should be long enough to provide adequate information for accurate decoding outcomes while maintaining the motor decoder's low complexity and high practicality. The deep learning AI motor decoder takes the 1 sec segment of the past data to predict motor intention at any point in time. Figure \ref{Fig_PastData}(A) shows analysis on the impact of input data length on the prediction accuracy, which is done with a subset of Subject CS's dataset. The results confirm a negative nonlinear correlation between the input data length and the prediction errors. The motor decoder reaches its optimal accuracy at 1 sec input data length. Doubling the input length to 2 sec does not necessarily improve the prediction errors, yet it requires much more computational power from the decoder. Interestingly, Figure \ref{Fig_PastData}(A) shows that wrist prediction is highly accurate regardless of the data length. It is consistent with our observation that nerve activities during the wrist pronation have exceptionally high SNR and distinctly appear on specific channels of the median nerve. Thus, the data pattern of the wrist pronation is very different from those of the finger flexes.

To help amputees achieve dexterous and intuitive control of prosthetic arms, besides high accuracy, the AI neural decoder must ensure low prediction latency while performing real-time decoding of motor intent. In contrast to the strong correlation between the input data length and the prediction error, there is no association between the former and the time latency. Figure \ref{Fig_PastData}(B) shows that the increase of the input data length from 0.2 sec to 2 sec causes no significant time latency on the hand movement decoding. Therefore, it justifies the use of 1 sec past data input trials.
\begin{figure*}[p]
\centering
\includegraphics[trim=0 0 0 0, clip=true, width=1\textwidth]{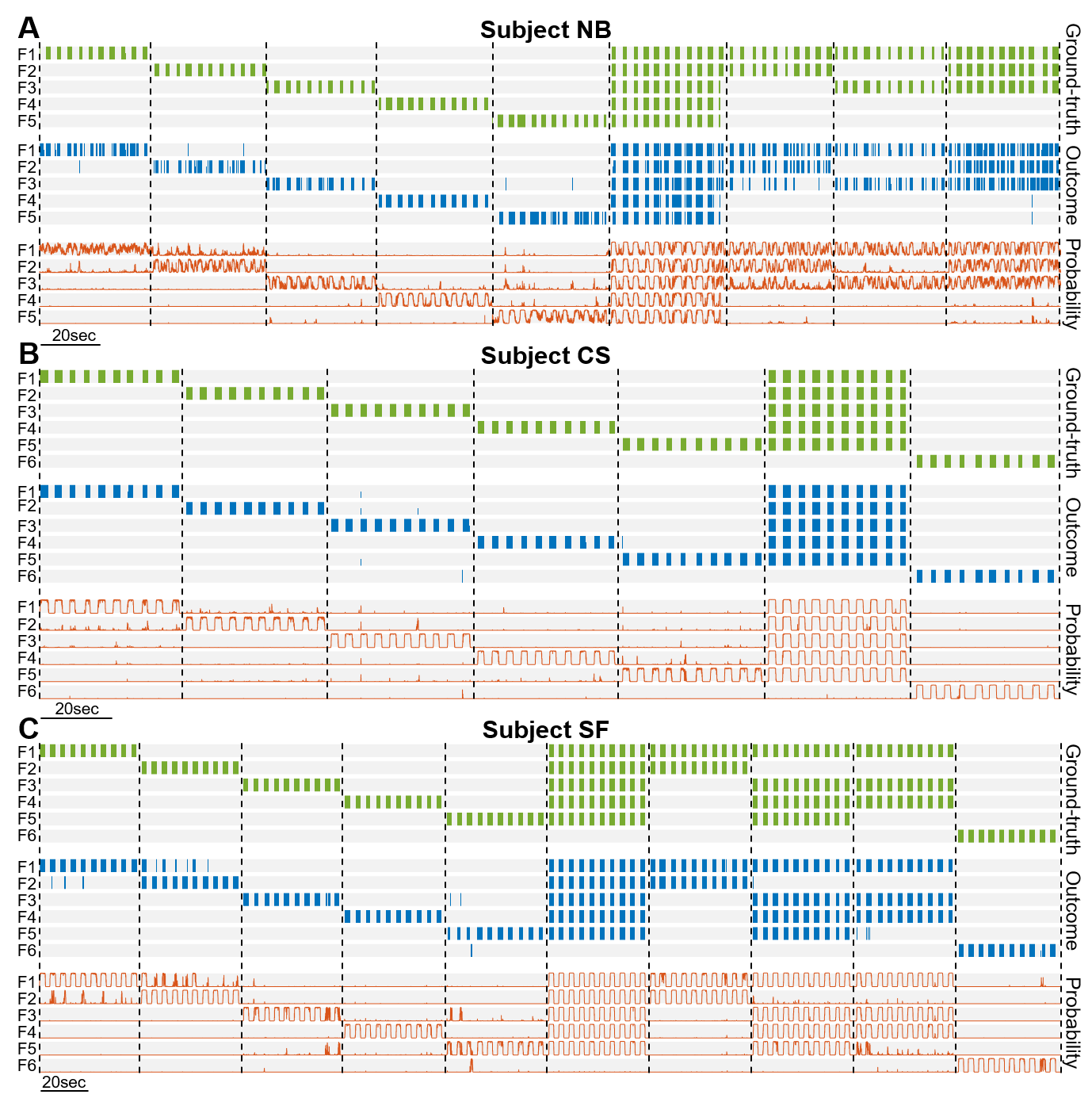}
\caption{(A, B, C) Classification results from motor decoding experiments with three subjects. The subjects repeat a hand gesture 10 times in each data segment. The AI neural decoder produces a prediction for 6 DOF every 20 msec using the past 1 sec of nerve data.}
\label{Fig_Predict}
\end{figure*}
\begin{table*}[ht]
\centering
\caption{\normalsize Summary of the classification performance metrics}
\includegraphics[width=1\textwidth]{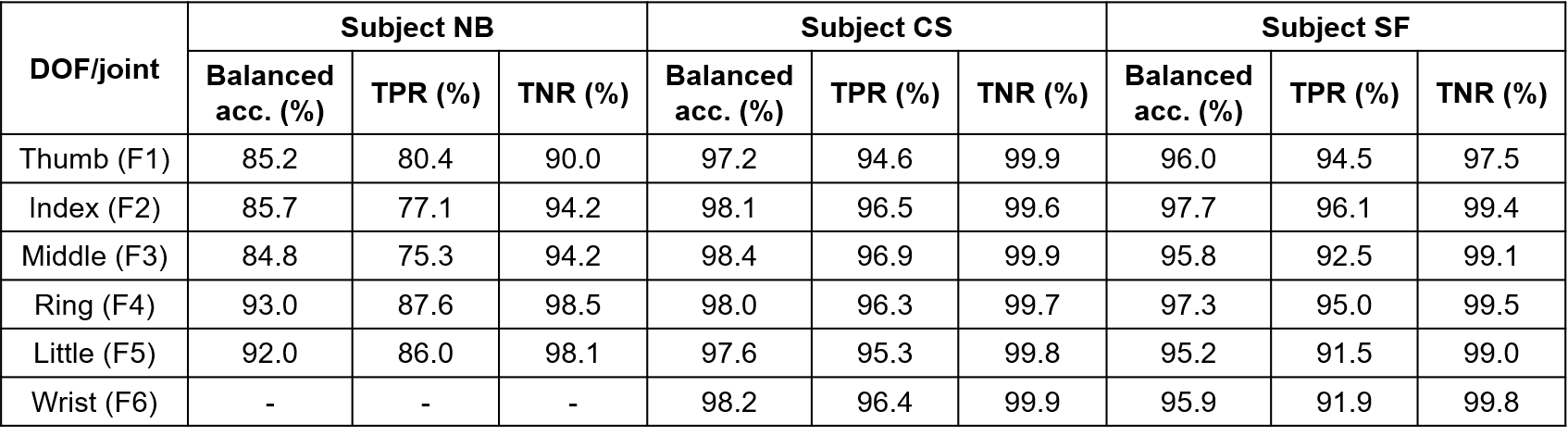}
\label{Table_Accuracy}
\end{table*}

Figure \ref{Fig_Predict} shows the motor intent classification outcomes of all three subjects. The subjects undergo several experiment sessions, each last a few days. The outcomes in Table \ref{Table_Accuracy} and Figure \ref{Fig_Predict} are from the last experiment session, in which training data are from the first day, and validation data are from the last day of the session. The subjects repeat a hand gesture 10 times in each data segment. However, each subject performs a slightly different set of hand gestures. They are individual finger flexing, fist, index pinch, middle pinch, and tripod pinch for Subject NB; individual finger flexing, fist, and wrist pronation for Subject CS; and lastly, individual finger flexing, fist, index pinch, pointing, horn, and wrist pronation, respectively in the order illustrated in Figure \ref{Fig_Predict} (A, B, C). Nevertheless, for comparison, they can be broken down into the individual performance of the 6 DOF.

Each of Figure \ref{Fig_Predict} (A, B, C) is equally divided into three parts. The top one-third of the rows are the ground truth of all DOF; the middle rows are the binary classification outcomes of each DOF after applying a threshold to the predicted probabilities in the last one-third rows. Subject NB was the first participant in this research and was not asked to perform the wrist pronation task. Data from Subject NB are recorded from only eight channels on the ulnar nerve. In addition, because Subject NB was explanted earlier than planned, we have less data for training and fine-tuning the deep learning architectures. This explains the classification outcomes of Table \ref{Table_Accuracy}.

The hand gesture classification outcomes from Subject NB are distinctly different for the first three and the last 2 DOF. The outcomes are significantly better for the ring and little fingers, which is consistent with the fact that his data are from the ulnar nerve only. However, the outcomes of the other three fingers are still good; the balanced accuracy, TPR, and TNR are up to 86.3\%, 78.7\%, and 94.2\%, respectively. Data from Subject CS and SF are recorded from 16 channels on both the median and ulnar nerve, resulting in more equivalent and accurate outcomes across all DOF. They are consistently above 90\% for all metrics. Specifically, the balanced accuracy, TPR, and TNR of Subject CS range from 97.2\% to 98.4\%, 94.6\% to 96.9\% and 99.6\% to 99.9\% in that order; those of Subject SF are 93.4\% to 97.9\%, 86.8\% to 96.1\%, and 97.3\% to 100\%. Although Subject CS shows slightly better decoding outcomes than Subject SF in terms of the three metrics above, Subject SF performs more hand gestures, including tricky ones such as index pinch, pointing, and horn. 

\subsection{Real-time and high-bandwidth transfer of motor information}
\begin{figure*}[p]
\centering
\includegraphics[trim=0 10 0 0, clip=true, width=0.9\textwidth]{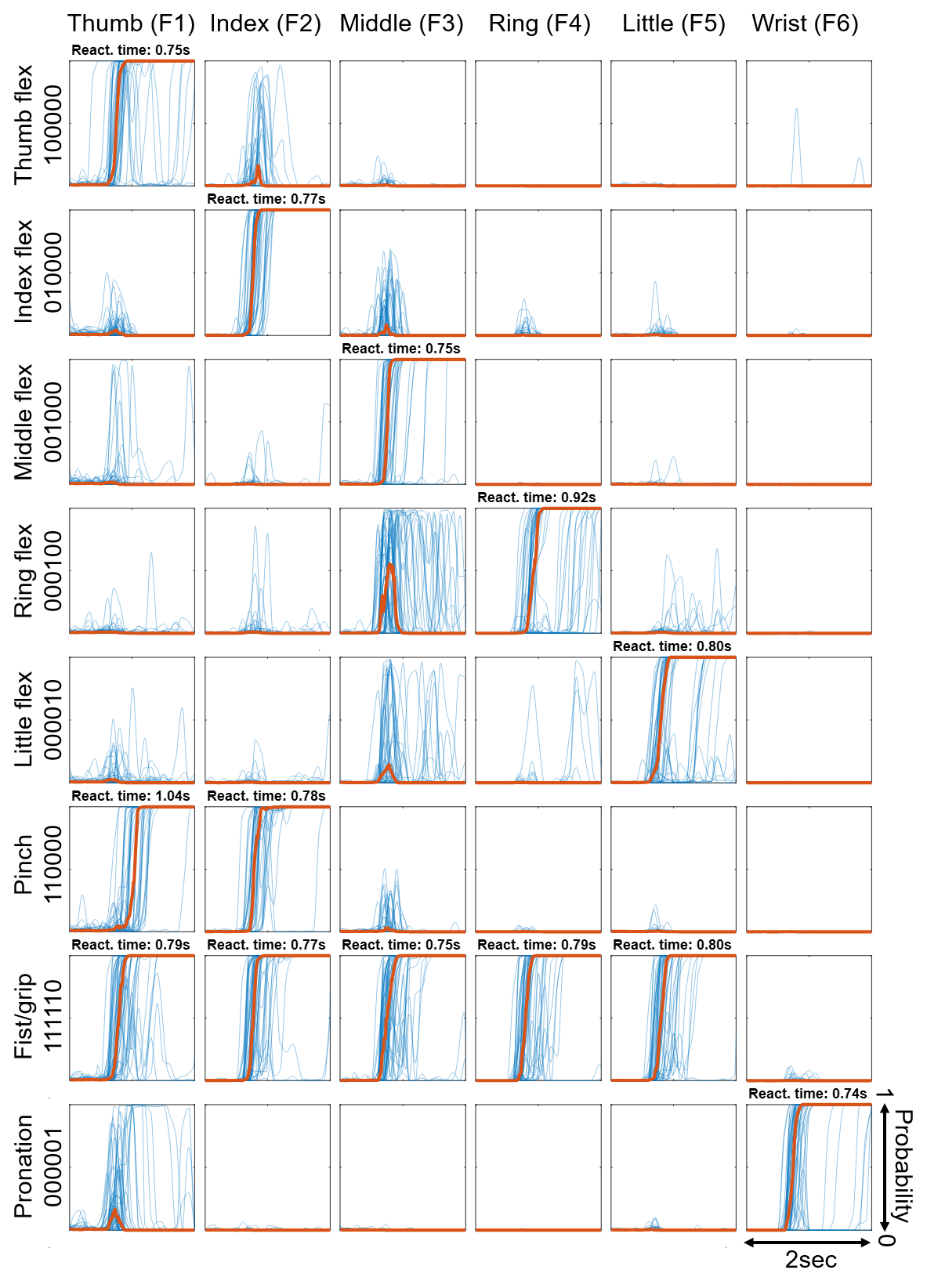}
\caption{Prediction probability of individual DOF and hand gesture in the matching task. In most of the trials, the AI neural decoder produces an accurate prediction in the subject's first attempt with a median reaction time of approximately 0.7-0.8 sec.}
\label{Fig_Trajectory}
\end{figure*}

After demonstrating that the trained AI agent can achieve high validation accuracy with low time latency on data recorded several days apart, we further test the AI agent’s efficiency on the real-time hand matching task described in Section \ref{Hand_Matching} to measure the end-to-end reaction time and information throughput of the nerve interface and motor decoder. 

Figure \ref{Fig_Trajectory} represents the predicted probability of each DOF gesture-wise in the time domain. The red line is the median predicted probability of each DOF across all trials at every specific amount of time since a targeted gesture is shown. As mentioned, a gesture is matched when all 6 DOF are correctly predicted. Therefore, the reaction time of every gesture is the maximum reaction time of all DOF. For example, after 1.04 sec, the majority of the pinch trials reach the target in which the thumb and index fingers already flex, and the others stay resting.   
\begin{table*}[ht]
\centering
\caption{\normalsize Summary of performance metrics of the matching task}
\includegraphics[width=1\textwidth]{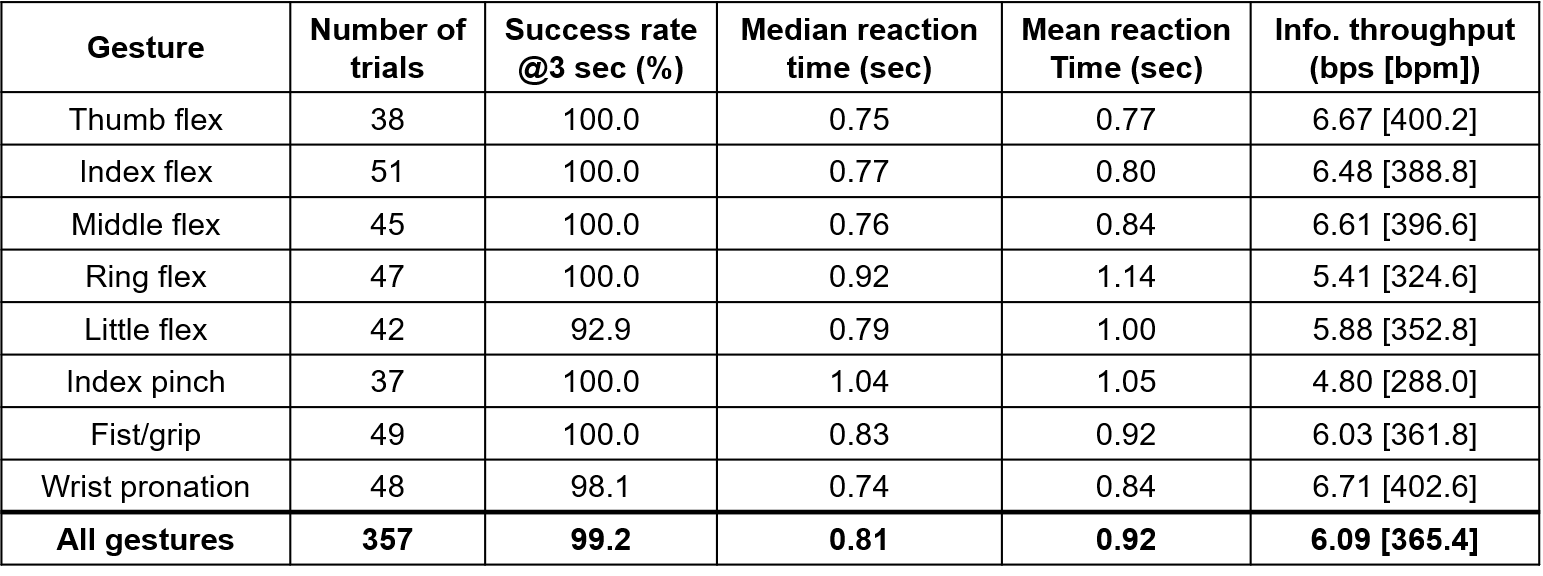}
\label{Table_Info}
\end{table*}

Table \ref{Table_Info} presents the success rate and the reaction time of all hand gestures, the two key factors to infer the information throughput given the information per trial is calculated above. The success rate, reaction time, and information throughput of all hand gestures are 99.2\%, 0.81 sec, and 6.09 bps (365.4 bpm), respectively. 
\begin{figure}[ht]
\centering
\includegraphics[trim=0 10 0 0, clip=true, width=0.6\textwidth]{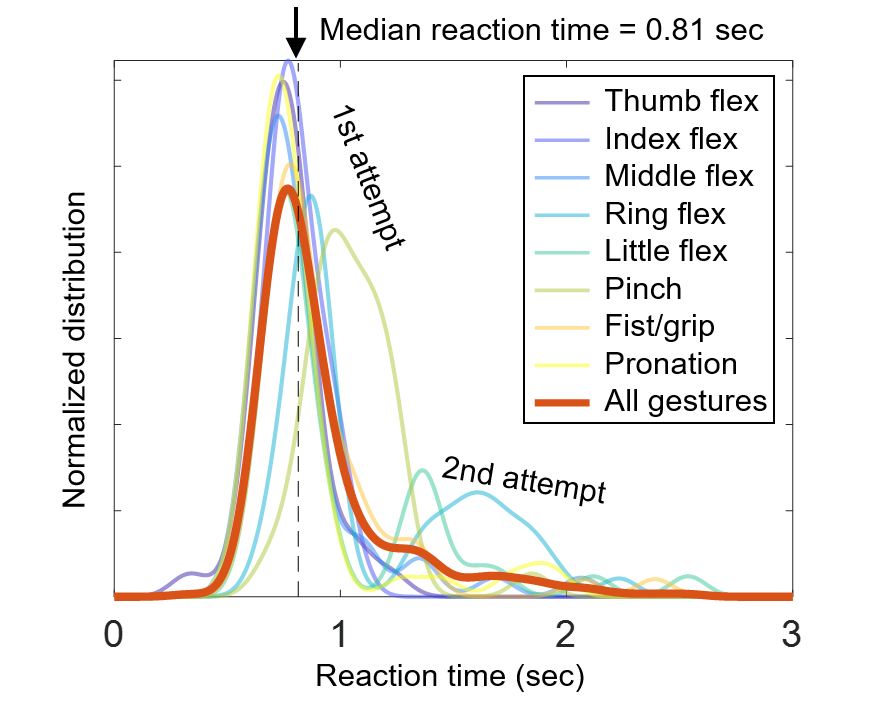}
\caption{Distribution of the reaction time across different gestures.}
\label{Fig_ReactionTime}
\end{figure}

Figure \ref{Fig_ReactionTime} shows the reaction time distribution across each hand gesture; the red line is a pooled distribution of all gestures. It is obtained by feeding the predicted probability of all DOF in all trials to the kernel density estimator. The results show that the AI agent accurately predicts Subject SF's first attempt in the majority of trials, represented by the distribution's primary peak at 0.7-0.8 sec. The secondary peak at 1.4-1.6 sec of the ring flex and the little flex represents the subject's second attempt. As shown in Figure \ref{Fig_Trajectory}, there are several false-positive predictions of the DOF F3, which is supposed to stay resting in both gestures. Furthermore, our analysis suggests that most of the time latency is from the human side of Equation \ref{Eq_ReactionTime} because for every real-time prediction outcome, the nerve data processing step takes less than 1 msec, and the motor decoding step requires only 10-20 msec. Nevertheless, this time latency is subject to the computer's CPU and GPU processing power. Running the decoder on an edge-computing device like the Jetson Nano as shown in \cite{2021_Nguyen_Draco} could have significantly higher time latency.

\subsection{Long-term deployment of a neuroprosthetic system}

For it to be suitable for practical applications, the AI agent should maintain robust performance over time. Therefore, we evaluate the agent stability by measuring its performance on Subject SF who has the longest implantation course of 16 months, in terms of signal strength, signal-to-noise ratio, reaction time, success rate, information throughput, and model persistence. 
\begin{figure*}[p]
\centering
\includegraphics[trim=0 0 0 0, clip=true, width=0.95\textwidth]{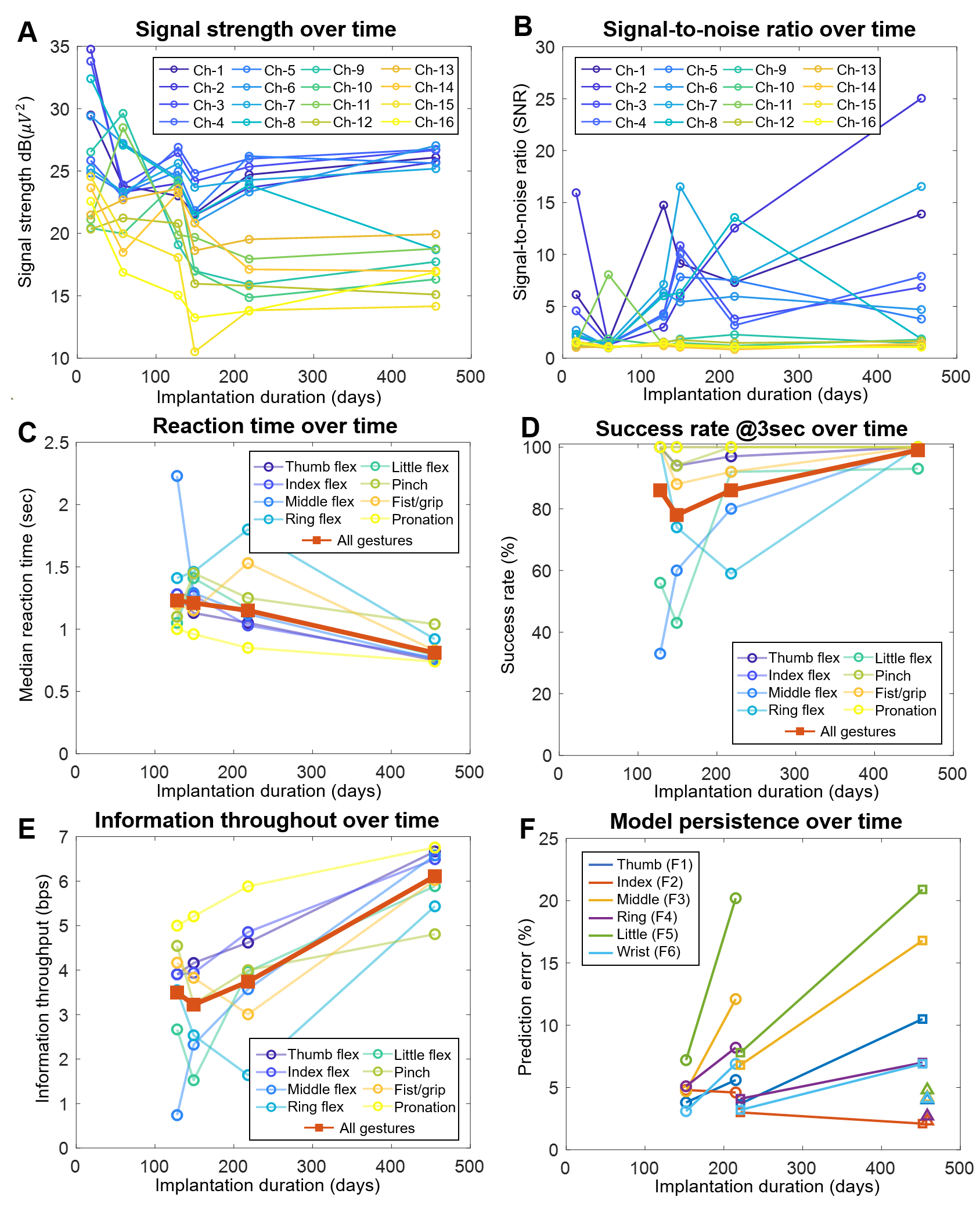}
\caption{(A, B) Changes in signal strength and SNR on different electrodes over the implantation duration. (C, D, E) Reaction time, success rate, and information throughput tested at different points over the implantation duration. (F) Evaluation of AI model persistence by training the network with the dataset from one experiment and validating with the dataset from another experiment a few months later.}
\label{Fig_Persistance}
\end{figure*}

Figure \ref{Fig_Persistance} (A, B) shows no trend of deterioration in the signal strength and SNR on 16 electrodes over the implantation duration. Moreover, thanks to the long implantation course of Subject SF, the authors have more time to fine-tune and retrain the AI model, and the subject has time to practice with the agent. Hence, there is a significant reduction in the reaction time from 1.23 sec to 0.81 sec and an improvement in the success rate from 86\% to 99.1\% for all gestures over time (Figure \ref{Fig_Persistance} (C, D)). In addition, the success rate converges, which leads to the more consistent and accurate performance of all fingers. This outcome is consistent with that of able people who also achieve a shorter reaction time with practice \cite{2006_Jensen}. Consequently, it causes an increase in the information throughput, which reaches 6.09 bps on the last experiment session before the explant surgery (Figure \ref{Fig_Persistance} (E)).

In addition, although the prediction error of the AI agent tends to go up, it is not an ever-increasing trend. As shown in Figure \ref{Fig_Persistance} (F), the prediction errors rise from 3.1\% to 7.2\% on day 152 to 4.6\% to 20.2\% on day 222 for 6 DOF, in which DOF F2 performs even better on day 222. Hence, the accuracy reduction is 13\% at most over 70 days. However, after the decoder is retrained, the prediction errors go down to 3.0\% to 7.8\%, equivalent to those of day 152. Furthermore, the reduction of the prediction accuracy on day 222 to day 459 is 13.1\% top, comparable to that of day 152 to day 222, considering the time gap is three times longer, which is 237 days. Retraining the AI model on the last experiment session brings the prediction errors down to 2.3\% to 4.8\%; those are even lower than the prediction errors of the first session. Thus, it reaffirms that there is no deterioration in the AI agent's long-term performance. In addition, if we set a threshold of 90\% accuracy, there are only DOF F4 and F5 underperforming after 70 days from day 152 to 222, which means the AI agent only needs to be fine-tuned every few months to maintain satisfying decoding outcomes.

\section{Discussion}
\label{Sec_Discussion}

\subsection{Hallmarks of robust prosthetic control}

The results demonstrate that employing an AI neural decoder through a nerve interface has hallmarks of robust prosthetic control. Numerous commercial prostheses use alternative muscle flexes as command inputs. However, such prostheses require more intensive user training and do not yield a natural user experience. The AI agent allows amputees to control prosthetic upper limbs with their thoughts by decoding true motor intent. Moreover, the agent can result in dexterous hand gestures by simultaneously decoding multiple DOF. Thus, with enough training data, the agent can take full advantage of near-anatomic prostheses like the LUKE Arm, opening the amputees up to a future of various hand movements, which is the necessary conditions for natural hand control. In addition, it needs to produce the desired movements with minimal reaction lag and high accuracy. The hand matching task proves that this agent can deliver satisfying real-time performance by producing highly accurate predicting outcomes of over 99\% for all gestures and low latency of about 0.81 sec, leading to a substantial amount of information throughput (365.4 bpm).

\subsection{Toward bidirectional nerve interface}

We aim to establish a neural interface that supports bidirectional communication between the human mind and computer. The AI neural decoder reported in this manuscript complements our previous works \cite{2019_Overstreet, 2021_Nguyen_RXF, 2021_Cheng}, in which we develop a neural stimulator and create distinct and reliable sensory feedback for amputees by neural stimulators. In particular, \cite{2019_Overstreet} shows a prior part of this clinical trial where the four amputees (three partial and one transradial) reported tactile and cutaneous sensations during stimulation of sensory fascicles and deeper proprioceptive sensations during stimulation of motor fascicles. These subjects experience different amputation lengths, ranging from 9 months to over 20 years. Nevertheless, the neural stimulators successfully enable sensory restoration from all subjects, including those with over 20 years of amputation. Putting together, this manuscript and our previous works form the foundation to materialize a complete closed-loop human-machine bidirectional communication.

However, many other factors need to be considered to achieve natural and accurate motor control of a prosthetic hand, including finger extension, wrist extension, wrist supination, hand movement directional prediction, and applied force estimation. By establishing a highly accurate AI neural decoder, some of these factors, such as finger and wrist extension, can be accounted for by extending the recorded datasets without needing a significant modification in the current motor decoding system architecture. However, tackling the other factors requires much change from our current system. Specifically, the wrist supination prediction task involves data recorded from the radial nerve, which requires additional microelectrode implants. In addition, to address the regression prediction of hand movements, including force and velocity estimation, we must redesign the experimental paradigm and use a different method to collect input data and the ground truth. These could be the new goal and addressed in our future works.

\subsection{Mind control beyond prostheses}
\begin{figure*}[ht]
\centering
\includegraphics[trim=0 0 0 0, clip=true, width=1\textwidth]{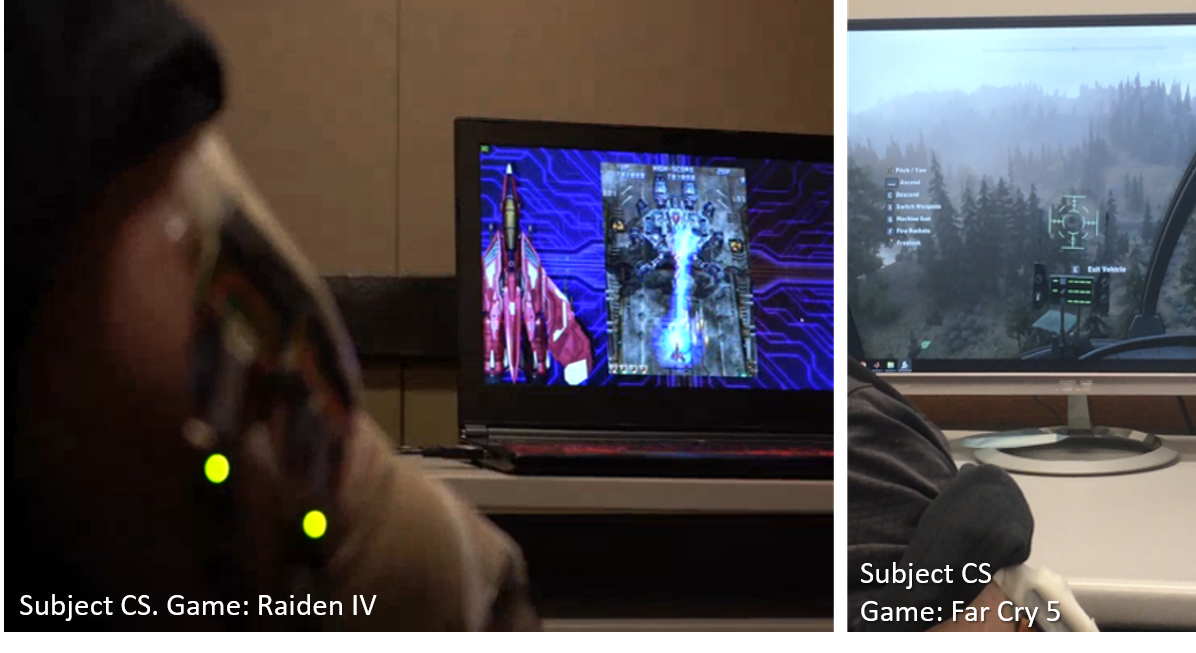}
\caption{We bind the hand gesture predictions to individual keystrokes on the computer, allowing Subject CS to play video games with only his thoughts. A similar setup can be used to control various devices and gadgets in a real or virtual environment.}
\label{Fig_Gaming}
\end{figure*}

A highly accurate AI neural decoder enables a future of various exciting applications. In addition to assisting amputees in their daily basic needs, such as intuitive control of prosthetic limbs, the AI agent can help with entertainment. In the experiment shown in Figure \ref{Fig_Gaming}, we bind the hand gesture predictions to individual keystrokes on the computer, allowing Subject CS to play video games with his thoughts. A similar setup can be used to control various devices and gadgets in a real or virtual environment. Instead of sending the predicted movement intent to actuate the prosthetic hand, the AI engine can wirelessly transmit the results to a remote controller to use a computer, virtual reality, fly a drone, control a robot, and so on. Furthermore, the users are not limited to amputees but anyone who receives the nerve interface implant. Hence, the proposed nerve interface with an AI neural decoder allows people to manipulate remote objects using only their thoughts in an actual ``telekinesis'' manner.

\section{Conclusion}
\label{Sec_Conclusion}

The purpose of this research is to prove the possibility of achieving intuitive real-time control and robust long-term performance of a prosthetic hand via a neuroprosthetic system using an AI agent to decode motor intent from the residual peripheral nerves. The AI neural decoder results in highly accurate outcomes in balanced accuracy, TPR, TNR from all three subjects in all 6 DOF, enabling the subjects to freely control their fingers and wrists by thoughts rather than via alternate muscle flexes. In addition, the hand matching test shows that the AI agent yields satisfying performance in real-time with a very high success rate and short reaction time, which leads to high information throughput across different gestures. Moreover, the neuroprosthetic system shows no sign of deterioration. Over 16 months, the signal strength and SNR from Subject SF are stable. The long implantation course and good signal quality allow the subject time to get used to the system. Hence, the reaction time gradually goes down, causing an increase in the information throughput. Furthermore, the decoding model performance is persistent over time. Without being retrained, the model prediction accuracy is still above 90\% for most DOF after 2 months and can go beyond 95\% if retrained even after 7 months. These results promise a future generation of prosthetic hands that can provide a natural user experience just like real hands.  

\section*{Declaration of Interest}

The surgery and patients related costs were supported in part by the DARPA under Grants HR0011-17-2-0060 and N66001-15-C-4016. The human motor decoding experiments, including the development of the prototype, was supported in part by MnDRIVE Program and Institute for Engineering in Medicine at the University of Minnesota, in part by the NIH under Grant R01-MH111413-01, in part by NSF CAREER Award No. 1845709.

Zhi Yang is co-founder of, and holds equity in, Fasikl Incorporated, a sponsor of this project. This interest has been reviewed and managed by the University of Minnesota in accordance with its Conflict of Interest policy. J. Cheng and E. W. Keefer have ownership in Nerves Incorporated, a sponsor of this project.


\begin{thebibliography}{999}

\bibitem{2008_ZieglerGraham} K. Ziegler-Graham \textit{et al.}, ``Estimating the prevalence of limb loss in the United States: 2005 to 2050,'' \emph{Archives of physical medicine and rehabilitation}, vol. 89, no. 3, pp. 422-429, 2008.

\bibitem{2012_Fougner} A. Fougner \textit{et al.}, ``Control of upper limb prostheses: Terminology and proportional myoelectric control-A review,'' \emph{IEEE Transactions on neural systems and rehabilitation engineering}, vol. 20, no. 5, pp. 663-677, 2012.

\bibitem{2012_Jiang} N. Jiang \textit{et al.}, ``EMG-based simultaneous and proportional estimation of wrist/hand kinematics in uni-lateral trans-radial amputees,'' \emph{Journal of neuroengineering and rehabilitation}, vol. 9, no. 1, pp. 1-11, 2012.

\bibitem{2013_Amsuss} S. Amsuss \textit{et al.}, ``Self-correcting pattern recognition system of surface EMG signals for upper limb prosthesis control,'' \emph{IEEE Transactions on Biomedical Engineering}, vol. 61, no. 4, pp. 1167-1176, 2013.

\bibitem{1960_Kobrinskiy} A. Kobrinskiy, ``Bioelectrical Control of Prosthetic Devices,'' \emph{Her Acad Sci}, vol. 30, no. pp. 58-61, 1960.

\bibitem{1965_Popov} B. Popov, ``The Bio-Electrically Controlled Prosthesis,'' \emph{The Journal of Bone and Joint Surgery. British Volume}, vol. 47, no. 3, pp. 421-424, 1965.

\bibitem{2012_Alkan} A. Alkan and M. Gunay, ``Identification of EMG signals using discriminant analysis and SVM classifier,'' \emph{Expert systems with Applications}, vol. 39, no. 1, pp. 44-47, 2012.

\bibitem{2015_Ortiz} M. Ortiz-Catalan \textit{et al.}, ``Offline accuracy: a potentially misleading metric in myoelectric pattern recognition for prosthetic control,'' in \emph{Proc. 2015 37th Annual International Conference of the IEEE Engineering in Medicine and Biology Society (EMBC)}, pp. 1140-1143, 2015.

\bibitem{2021_Souza} J. O. d. O. de Souza \textit{et al.}, ``Investigation of Different Approaches to Real-Time Control of Prosthetic Hands With Electromyography Signals,'' \emph{IEEE Sensors Journal}, vol. 21, no. 18, pp. 20674-20684, 2021.

\bibitem{2016_Cordella} F. Cordella \textit{et al.}, ``Literature review on needs of upper limb prosthesis users,'' \emph{Frontiers in neuroscience}, vol. 10, pp. 209, 2016.

\bibitem{2010_Schalk} G. Schalk, ``Can electrocorticography (ECoG) support robust and powerful brain-computer interfaces?,'' \emph{Frontiers in neuroengineering}, vol. 3, no. pp. 9, 2010.

\bibitem{2016_Flesher} S. N. Flesher \textit{et al.}, ``Intracortical microstimulation of human somatosensory cortex,'' \emph{Science translational medicine}, vol. 8, no. 361, pp. 361ra141-361ra141, 2016.

\bibitem{2018_Collinger} J. L. Collinger \textit{et al.}, ``Progress towards restoring upper limb movement and sensation through intracortical brain-computer interfaces,'' \emph{Current Opinion in Biomedical Engineering}, vol. 8, no. pp. 84-92, 2018.

\bibitem{2018_Xu_Nerve} J. Xu \textit{et al.}, ``A Bidirectional Neuromodulation Technology for Nerve Recording and Stimulation,'' \emph{Micromachines}, vol. 9, no. 11, pp. 538, 2018.

\bibitem{2020_Nguyen_Scorpius} A. T. Nguyen \& J. Xu \textit{et al.}, ``A Bioelectric Neural Interface Towards Intuitive Prosthetic Control For Amputees,'' \emph{Journal of Neural Engineering}, vol. 17, no. 6, pp. 066001, 2020.

\bibitem{2021_Nguyen_RXF} A. T. Nguyen \textit{et al.}, ``Redundant Crossfire: A Technique to Achieve Super-Resolution in Neurostimulator Design by Exploiting Transistor Mismatch,'' \emph{IEEE Journal of Solid-State Circuits}, vol. 56, no. 8, pp. 2452-2465, 2021.

\bibitem{2015_Pasquina} P. F. Pasquina \textit{et al.}, ``First-in-man demonstration of a fully implanted myoelectric sensors system to control an advanced electromechanical prosthetic hand,'' \emph{Journal of neuroscience methods}, vol. 244, no. pp. 85-93, 2015.

\bibitem{2015_Saal} H. P. Saal \& S. J. Bensmaia, ``Biomimetic approaches to bionic touch through a peripheral nerve interface,'' \emph{Neuropsychologia}, vol. 79, pp. 344-353, 2015.

\bibitem{2015_Schiefer} M. Schiefer \textit{et al.}, ``Sensory feedback by peripheral nerve stimulation improves task performance in individuals with upper limb loss using a myoelectric prosthesis,'' \emph{Journal of neural engineering}, vol. 13, no. 1, pp. 016001, 2015.

\bibitem{2016_Davis} T. S. Davis \textit{et al.}, ``Restoring motor control and sensory feedback in people with upper extremity amputations using arrays of 96 microelectrodes implanted in the median and ulnar nerves,'' \emph{Journal of neural engineering}, vol. 13, no. 3, pp. 036001, 2016.

\bibitem{2016_Lachapelle} J. R. Lachapelle \textit{et al.}, ``An Implantable, Designed-for-Human-Use Peripheral Nerve Stimulation and Recording System for Advanced Prosthetics,'' in \emph{Proc. Annual International Conference of the IEEE Engineering in Medicine and Biology Society (EMBC)}, pp. 1794-1797, 2016.

\bibitem{2017_Vu} P. P. Vu \textit{et al.}, ``Closed-Loop Continuous Hand Control via Chronic Recording of Regenerative Peripheral Nerve Interfaces,'' \emph{IEEE Transactions on Neural Systems and Rehabilitation Engineering}, vol. 26, no. 2, pp. 515-526, 2017.

\bibitem{2017_Wendelken} S. Wendelken \textit{et al.}, ``Restoration of Motor Control and Proprioceptive and Cutaneous Sensation in Humans with Prior Upper-Limb Amputation via Multiple Utah Slanted Electrode Arrays (USEAs) Implanted in Residual Peripheral Arm Nerves,'' \emph{Journal of Neuroengineering and Rehabilitation}, vol. 14, no. 1, pp. 121, 2017.

\bibitem{2019_Wolf} E. J. Wolf \textit{et al.}, ``Advanced Technologies for Intuitive Control and Sensation of Prosthetics,'' \emph{Biomedical Engineering Letters}, vol. 10, pp. 119-128, 2019.

\bibitem{2020_Cracchiolo} M. Cracchiolo \textit{et al.}, ``Decoding of Grasping Tasks from Intraneural Recordings in Trans-Radial Amputee,'' \emph{Journal of Neural Engineering}, vol. 17, no. 2, pp. 026034, 2020.

\bibitem{2020_Vu} P. P. Vu \textit{et al.}, ``A Regenerative Peripheral Nerve Interface Allows Real-Time Control of an Artificial Hand in Upper Limb Amputees,'' \emph{Science Translational Medicine}, vol. 12, no. 533, pp. eaay2857, 2020.

\bibitem{2014a_Resnik} L. Resnik \textit{et al.}, ``The DEKA Arm: Its Features, Functionality, and Evolution During the Veterans Affairs Study to Optimize the DEKA Arm,'' \emph{Prosthetics and Orthotics International}, vol. 38, no. 6, pp. 492-504, 2014.

\bibitem{2014b_Resnik} L. Resnik \textit{et al.}, ``User and Clinician Perspectives on DEKA Arm: Results of VA Study to Optimize DEKA Arm,'' \emph{Journal of Rehabilitation Research and Development}, vol. 51, no. 1, pp. 27-38, 2014.

\bibitem{2012_Sussillo} D. Sussillo \textit{et al.}, ``A Recurrent Neural Network for Closed-Loop Intracortical Brain–Machine Interface Decoders,'' \emph{Journal of Neural Engineering}, vol. 9, no. 2, pp. 026027, 2012.

\bibitem{2016_Atzori} M. Atzori \textit{et al.}, ``Deep Learning with Convolutional Neural Networks Applied to Electromyography Data: A Resource for the Classification of Movements for Prosthetic Hands,'' \emph{Frontiers in Neurorobotics}, vol. 10, pp. 9, 2016.

\bibitem{2018_George} J. A. George \textit{et al.}, ``Improved Training Paradigms and Motor-Decode Algorithms: Results from Intact Individuals and A Recent Transradial Amputee with Prior Complex Regional Pain Syndrome,'' in \emph{Proc. Annual International Conference of the IEEE Engineering in Medicine and Biology Society (EMBC)}, pp. 3782-3787, 2018.

\bibitem{2019_Alazrai} R. Alazrai \textit{et al.}, ``A Deep Learning Framework for Decoding Motor Imagery Tasks of the Same Hand Using EEG Signals,'' \emph{IEEE Access}, vol. 7, pp. 109612-109627, 2019.

\bibitem{2019_Dantas} H. Dantas \textit{et al.}, ``Deep Learning Movement Intent Decoders Trained with Dataset Aggregation for Prosthetic Limb Control,'' \emph{IEEE Transactions on Biomedical Engineering}, vol. 66, no. 11, pp. 3192-3203, 2019.

\bibitem{2020b_George} J. A. George \textit{et al.}, ``Inexpensive and Portable System for Dexterous High-Density Myoelectric Control of Multiarticulate Prostheses,'' in \emph{Proc. IEEE International Conference on Systems, Man and Cybernetics (SMC)}, pp. 3441-3446, 2020.

\bibitem{2021_Luu_Frontiers} D. K. Luu \textit{et al.}, ``Deep Learning-Based Approaches for Decoding Motor Intent from Peripheral Nerve Signals,'' \emph{Frontiers in Neuroscience}, vol. 15, pp. 667907, 2021.

\bibitem{2021_Nguyen_Draco} A. T. Nguyen \& M. W. Dealan \textit{et al.}, ``A Portable, Self-Contained Neuroprosthetic Hand with Deep Learning-Based Finger Control,'' \emph{Journal of Neural Engineering}, vol. 18, no. 5, pp. 056051, 2021.

\bibitem{1978_Posner} M. I. Posner, \emph{Chronometric explorations of mind}. Lawrence Erlbaum, 1978. 

\bibitem{2006_Jensen} A. R. Jensen, \emph{Clocking the mind: Mental chronometry and individual differences}. Elsevier, 2006.

\bibitem{2017_Cheng} J. Cheng \textit{et al.}, ``Dexterous Hand Control Through Fascicular Targeting (HAPTIX-DEFT): Level 4 Evidence,'' \emph{Journal of Hand Surgery}, vol. 42, no. 9, pp. S8-S9, 2017.

\bibitem{2019_Overstreet} C. K. Overstreet \textit{et al.}, ``Fascicle Specific Targeting For Selective Peripheral Nerve Stimulation,'' \emph{Journal of Neural Engineering}, vol. 16, no. 6, pp. 066040, 2019.

\bibitem{2021_Cheng} J. Cheng \textit{et al.}, ``Fascicle-Specific Targeting of Longitudinal Intrafascicular Electrodes for Motor and Sensory Restoration in Upper-Limb Amputees,'' \emph{Hand Clinics}, vol. 37, no. 3, pp. 401-414, 2021.

\bibitem{1948_Shannon} C. E. Shannon, ``A Mathematical Theory of Communication,'' \emph{The Bell System Technical Journal}, vol. 27, no. 3, pp. 379-423, 1948.

\bibitem{2014_Kingma} D. P. Kingma \& J. Ba, ``Adam: A Method for Stochastic Optimization,'' \emph{arXiv}, 1412.6980, 2014.

\end{thebibliography}
\end{document}